\documentclass[letterpaper,10pt,twocolumn]{article}

\usepackage{fancyhdr}
\usepackage[margin=1.0in]{geometry}

\usepackage{amsmath, amssymb, amsfonts}
\usepackage{euscript}

\usepackage{url}
\usepackage{natbib}

\usepackage{verbatim}

\newtheorem{definition}{Definition}

\newtheorem{remark}{Remark}

\newenvironment{tightitemize}
{ \begin{itemize}
    \setlength{\itemsep}{0pt}
    \setlength{\parskip}{0pt}
    \setlength{\parsep}{0pt}     }
{ \end{itemize}                  } 

\title{Natively Interpretable Machine Learning and Artificial Intelligence: Preliminary Results and Future Directions}

\author{Christopher J. Hazard\thanks{Diveplane Corporation.  If you are interested in using our technology, please contact \texttt{info@diveplane.com}.  The authors would like to thank the investors, employees, and supporters of Diveplane Corporation for making this work possible.}, Christopher Fusting\footnotemark[1], Michael Resnick\footnotemark[1], Michael Auerbach\footnotemark[1],\\Michael Meehan\footnotemark[1], Valeri Korobov\footnotemark[1]}

\begin{document} 

\maketitle
\thispagestyle{fancy}

\section{Abstract}
Machine learning models have become more and more complex in order to better approximate complex functions. Although fruitful in many domains, the added complexity has come at the cost of model interpretability. The once popular k-nearest neighbors (kNN) approach, which finds and uses the most similar data for reasoning, has received much less attention in recent decades due to numerous problems when compared to other techniques.  We show that many of these historical problems with kNN can be overcome, and our contribution has applications not only in machine learning but also in online learning, data synthesis, anomaly detection, model compression, and reinforcement learning, without sacrificing interpretability.  We introduce a synthesis between kNN and information theory that we hope will provide a clear path towards models that are innately interpretable and auditable.  Through this work we hope to gather interest in combining kNN with information theory as a promising path to fully auditable machine learning and artificial intelligence.

\section{Introduction and Motivations}

As machine learning has matured the need to understand, interpret and explain models has become increasingly important~\citep{AlpaydinBook, MohriFoundationsBook, GoodfellowBook}. Machine learning models are interpreted in a variety of ways including exploring the internals of a model~\citep{SkapuraBook, poerner2018evaluating}, creating ex post rationalizations~\citep{ribeiro2016lime, google2018whatif} or using models that are interpretable from the beginning and maximize their accuracy~\citep{wang2015falling}. There is a perception and some supporting evidence that there exists a trade off between accuracy and interpretability~\citep{EvolutionaryTraining}.

The motivating philosophy behind our work is that models should be innately interpretable. Specifically, our motivations are:
\begin{tightitemize}
	\item decisions should be directly traceable to the training data that caused the decision to be made;
	\item the regions of the model should be easily characterized directly from the training data; and
	\item assumptions are minimal.
\end{tightitemize}
To achieve the aforementioned goals we combine k-Nearest Neighbors (kNN) with the principle of maximum entropy to create models that are easy to understand, make minimal assumptions, and are nonparametric. 

k-Nearest Neighbors is one of the oldest, simplest, and most accurate algorithms for pattern classification and regression models ~\citep{HastieStatisticalBook}. It is a simple technique that is easily implementable ~\citep{AlpaydinVoting}. The accuracy of kNN-based classification, prediction, and recommendation depends solely on a data model. Outputs from the model are usually traceable back to the exact data that influenced each decision.  This traceability enables detailed analysis of the decision inputs and characterization of the data local to the decision.

k-Nearest Neighbors was previously a dominant machine learning technology~\citep{coomansAlternative, BreimanTree, AltmanIntroduction, AlpaydinVoting} but was largely abandoned with the growing size of data and the computational complexity of finding the nearest $k$ points~\citep{Raikwalperformance, SchuhMitigating, ArabicArticles2008}. Many optimizations have been proposed over the years, they generally seek to reduce the number of distances actually computed  \citep{pedregosa2011scikitlearn}. The optimizations include linear scan, Kd-trees, balltrees, etc. \citep{pedregosa2011scikitlearn}. The curse of dimensionality has also been known to adversely affect kNN \citep{HastieStatisticalBook, IndykApprox, SchuhImproving, TaoQuality} and the selection of a distance function can be challenging \citep{SuryaDist}. Additionally features may have to be scaled or standardized to prevent distance measures from being dominated by one of the features. The accuracy of kNN can be severely degraded by the presence of noisy or irrelevant features, or if the features scales are not consistent with their relevance. Finally, kNN requires a value of parameter k. If k is too small, the model may have low bias but be sensitive to noisy points and have too high of variance. If k is too large, the neighborhood may include points from other classes and may have too little variance.

Our contributions in this paper are several. First, we bring numerous well-studied techniques together to improve the efficacy of kNN. Second, we connect kNN with information theory and describe numerous ways this can be applied to machine learning. Third, we illustrate how the first two contributions open the door to interpretable reinforcement learning. Throughout this paper we discuss targetless models (models in which we are interested in predicting all the features using all the others), their utility in understanding the data, and introduce an imputation method that naturally arises from such models.

The purpose of this paper is to offer a glimpse as to what the combination of kNN and information theory can offer in advancing the state of the art of machine learning and artificial intelligence.

\section{Targetless kNN and Entropy}
\label{sec:targetless_knn_entropy}

We introduce term \emph{targetless learning} to describe our approach to kNN.  Instead of the traditional approach of building a model that learns the mapping from a set of input features variables to a set of target variables, or building multiple models to learn multiple mappings, our models consist of the relevant training data stored in a data structure that can be quickly queried.  We may wish to predict and characterize any set of variables given any other set of variables.  This flexibility is generally not emphasized in the related literature outside of a subset of work on semi-supervised learning and imputation \citep{tan2018incomplete,zhao2015semi}, and so we define two more terms to help us describe inputs and outputs in targetless learning.  \emph{Context features} are the feature variables being used as inputs for a particular query.  \emph{Action features} are the feature variables that are being labeled, actioned upon, or predicted; in traditional targeted machine learning, these are usually referred to as target variables, labels, or responses with regard to a targeted machine learning model.  These terms further reflect the origin and potential for online learning applications of this approach.

\subsection{Similarity of Points}
\label{ssec:sim_metric}

When determining the value of an unknown action feature, the action features of the $k$ most similar points are averaged or their values voted upon to determine the most likely value. In general similarity is determined by a distance metric. Unfortunately as the number of dimensions increases it becomes difficult to distinguish points \citep{hinneburg2000nearest, beyer1999nearest}. One proposed solution to this problem is to use the number of shared nearest neighbors as a similarity measure \citep{houle2010shared}. There is also evidence that fractional norms heading towards zero enable points to be distinguished more easily in high dimensional space \citep{aggarwal2001surprising}.  Fractional norms are represented as $||x||_p$ as
\begin{equation}
 \label{eq:frac_norm}
||x||_p = \left(\sum_{i \in \Xi} w_i x_{i}^p \right)^{1/p},
\end{equation}
where $p$ is the parameter for the Lebesgue space, $\Xi$ is the feature set, and $w_i$ is the weight for each feature, often where $w_i = \frac{1}{n}$.

Motivated by this result we derived the Minkowski distance as $p\to 0$ expressed over feature set $\Xi$
\begin{equation}
\label{eq:minkowski_distance_p0}
\lim_{p \to 0} d_p\left(x, y\right) = \left( \prod_{i \in \Xi} |x_i - y_i| \right)^{\frac{1}{|\Xi|}}.
\end{equation}
When feature weights $w_i$ satisfy $\sum_{i \in \Xi} w_i = 1$ we have
\begin{equation}
\label{eq:minkowski_distance_p0_weighted}
\lim_{p \to 0} d_p\left(x, y\right) = \prod_{i \in \Xi} |x_i - y_i|^{w_i}.
\end{equation}

Note equations \ref{eq:minkowski_distance_p0} and \ref{eq:minkowski_distance_p0_weighted} are geometric means and have the useful property of being scale invariant. Scale invariance means that scaling a feature by any factor will not affect the ordering of proximity, as the result is the same as multiplying all of the distances by a constant.  Thus using $p = 0$ enables the data to be stored in its original form, not scaled, standardized, or normalized, which improves the transparency of the model and removes the need for that aspect of feature engineering.  As $\lim p \to 0$, the scale of features matters less, meaning that the Minkowski distance is approximately scale invariant with regard to $p$ values that are relatively close to $0$.

\subsubsection{A Probabilistic Approach}

\label{sssec:probabilistic_distance}
%***TODO: relate to two-sample and hypothesis testing

We are unaware of any prior work that has investigated using $p = 0$ as a distance function.\footnote{Using $p = 0$ is in not a metric, and arguably not a distance function, as it fails the triangle inequality. Further, it is not technically $p = 0$ but rather $\lim_{p \to 0}$, but we use this abuse of notation for simplicity.} This is unsurprising, as $p = 0$ causes significant problems for any data set that has categorical data or other data that may be exactly equal.  Consider a data set that has two features.  If two points have equal values for the first feature, then the distance between the points will be zero regardless of the distance between the other values of the other feature, due to the multiplication by zero.

Instead, consider distance probabilistically, in the sense that each feature ``distance'' is the probability that the two values are different given the observations or measurements of the feature values (instead of being the absolute value of their difference) as a way of handling uncertainty \citep{agarwal2016nearest}.  If we assume that each observation is independent, then simply multiplying the probabilities that each feature value is the same yields the same distance measure as Equation~\ref{eq:minkowski_distance_p0} when determining the conjunction of the probabilities.  Solving the problems that come from exact matches using $p=0$ means we can work directly with probabilities or deal with features on wildly different scales without having to standardize or otherwise scale them.  Using the geometric mean to combine measurements of achieving different goals has been shown to be an effective objective function for multicriteria optimization \citep{harrington1965desirability}, and so using it to provide contrast between different similarities is a natural use.\footnote{We note that this realization was inspired by early blog post drafts of work done by \cite{leinster2016maximizing} in that the generalized diversity index, which can be parameterized to measure the Shannon entropy, is nothing more than the reciprocal of the generalized mean when substituting $p-1$ for $p$ when dealing with probabilities~\citep{tuomisto2010consistent}, and the Minkowski distance is just the generalized mean of the differences.}

Suppose we have made two observations of a value, $x$ and $y$ respectively, and we would like to know the ``distance'' between them.  The obvious distance of $|x - y|$ yields the maximum likelihood value of the distance, but does not yield the expected value of the distance.  Consider that there is considerable deviation among observations of the same value, meaning that there is likely to be a relatively large expected difference between observations of the same value.  We use the term deviation to encompass both the error, which pertains the difference between actual and measurement, and the residual, which pertains to the difference between actual and estimated.  This generic use of deviation applies regardless of whether the observation is measured, predicted, or inferred, and regardless of whether the deviation is due to randomness or lack of additional information that would reduce the deviation.

Consider two observations, $x$ and $y$, with considerable deviation.  If $x = 100 \pm 10$ and $y = 100 \pm 10$, intuitively the expected distance between $x$ and $y$ is likely to be greater than 0 even though the expected values is the same, yet a simple subtraction yields $0$.  Further, consider that $x$ and $y$ are feature vectors of length two of $x = \{1.1, 100\}$ and $y = \{1.2, 10\}$.  If we have a third observation that is $z = \{1.1, 10.01\}$, using $p = 0$ for measuring the similarity between $z$ and both $x$ and $y$ will yield $x$ as infinitely closer than $y$ because the difference between the first terms is zero and the multiplication makes the distance zero.  Though this may sometimes be desirable, larger deviations for the first feature and smaller deviations for the second feature should yield $y$ as more similar to $z$ than $x$.

To solve the problem of zero expected distance for identical measurements despite deviation, and to address the high sensitivity of $L_p$ with close or exact matches with a low $p$ value,  we employ the {\L}ukaszyk-Karmowski metric \citep{lukaszyk2003metryka,lukaszyk2004lkmetric}.  Given a probability density function of $x$, $f$, and a probability density function of $y$, $g$, the expected difference between them becomes
\begin{equation}
d(x, y) = \int_{-\infty}^\infty \int_{-\infty}^\infty |x-y| f(x) g(y) \, dx\, dy.
\end{equation}
We assume that if both points are near enough to be worth determining the distance between them, then the distributions and parameters for the probability density functions should represent the local data.  The two simple maximum entropy distributions on $(-\infty, \infty)$ given a point and a distance around the point are the Laplace distribution (double exponential) and Gaussian distribution, depending on whether the distance is represented as mean absolute error or standard deviation respectively.  The Gaussian or normal distribution has a clean closed form solution.  Letting $\mu_{xy} \equiv |x-y|$, the expected distance for two normal identical distributions becomes
\begin{equation}
d_{NN}(x, y) = \mu_{xy} + \frac{2\sigma}{\sqrt\pi}\operatorname{exp}\left(-\frac{\mu_{xy}^2}{4\sigma^2}\right)-\mu_{xy} \operatorname{erfc} \left(\frac{\mu_{xy}}{2\sigma}\right).
\end{equation}
For the previous example of $x = 100 \pm 10$ and $y = 100 \pm 10$, the expected distance is approximately $11.3$, which is more reasonable for two values that have a standard deviation of $10$.

In order to employ this measure, however, we need a value for the deviations.  Measurement error may not always be readily available, and it does not take into account the additional error among the relationships within the model.  For continuous values we can find the smallest nonzero distance between any two points in the data with regard to the feature.  Because that is the smallest observed distance, we do not know whether the model's relationships can yield a finer resolution, so this acts as an empirical lower bound for the deviations and a good starting value.

%***TODO: consider adding pseudocode for residual calculations
Alternatively residuals can be calculated for each prediction.  The mean absolute error or standard deviation can be calculated for each observation using a hold-one-out approach, where instances are removed from the model and each of the held out instance's features are predicted using the rest of the data.  These errors can be locally aggregated or can be aggregated across the entire model to obtain the expected residual, $r$, for predicting each feature, $i$, as $r_i$.  We have found that using the residuals in the kNN system with the {\L}ukaszyk-Karmowski metric, calculating new residuals, and then feeding these back in, generally yields convergence of the residual values with notable convergence after only 3 or 4 iterations.

Measuring a distance value for each feature further enables parameterization regarding the type of data a feature holds. For example, nominal data can result in a distance of 1 if the values are not equal and 0 if they are equal.  Thus, one-hot encoding, the expansion of nominal values into multiple features, is not needed.  Ordinal data can use a distance of 1 between each ordinal type.  Cyclical data can perform appropriate subtractions while keeping the data on a single dimension, which keeps the feature in one dimension and directly understandable rather than having to split the feature into two using trigonometry.

\subsection{Information in the Data}
We now quantify the amount of information in a kNN model. Because our formulation of kNN uses a similarity measure based on distance, we first quantify each point $x$ by the amount of distance it contributes to the $k$ nearest points. In general, we define \emph{conviction} as a normalized measure of how much \emph{surprisal} one would expect for a given situation relative to the surprisal observed.  If we have some form of prior distribution of data given all of the information observed up to that point, the surprisal is the amount of information gained when we observe a new sample, event, case, or state change and update the prior distribution to form a new posterior distribution after the event.  The surprisal of an event of observing a random variable $x \sim X$ is defined as $I(x) = -\ln p(x)$.  Thus, the conviction, $\pi$, can be expressed as
\begin{equation}
	\pi(x) = \frac{\mathbb{E}[I(X)]}{I(x)}.
\end{equation}

This ratio results in conviction values $\pi \in [0,\infty)$, where
\begin{tightitemize}
    \item $\pi = 0$ means this point has an infinite amount of surprisal, that is, the point was previously thought to be impossible to exist within the dataset;
    \item $\pi = 1$ means this point has an average amount of surprisal, that is, it adds an average amount of information to the model; and
    \item $\pi = \infty$ means this point is not at all surprising, that is, it is so redundant that the point could be discarded without affecting the model at all.
\end{tightitemize}

This ratio is indicative of how much information is required to encode one aspect of the model relative to another, whether dealing with cases or features.  In some cases conviction can act as a proxy for several matters, such as how confident we are about our data, whether the data is correct or anomalous, whether the data belongs together, or whether the data is useful in making predictions.  In other cases, conviction can inform how the model will be harmed if data is removed, and additionally can be used to control the surprisal when performing data synthesis.

Conviction can be computed in a targeted or targetless manner.  In a targeted manner, each feature or case is compared against another set of target cases or features one on one.  In an untargeted manner, each case or feature is held out one by one and compared against the rest of the data in the model.  When holding one out, the change in probability impact on other elements of the model indicates a measure of hubness or centrality of the data which, when isolated, has been found to be of significant importance for determining the influence of data on the model \citep{tomavsev2014hubness}.

If the probability space over which conviction is normalized is broadened (or if surprisal is used without normalization), then even the model impact of combinations of features can be compared to that of combinations of observations.

In the following sub-sections, we discuss different forms of conviction that can be derived.  More forms of surprisal and conviction can be conditioned and computed, opening up a rich area for different kinds of informativeness about various aspects of the model.

We note that conviction is related to, but not exactly the same as, feature importance or case influence and so care must be taken when comparing the two.\footnote{It is possible to compare the results of estimating values for a feature, and then compute a conviction ratio comparing the mutual information among a set of estimated values as an information theoretic representation of mean decrease in accuracy or Shapley value.  This would be an information theoretic result that is much closer to feature importance.}
%  We will discuss this further in Section~\ref{sssec:conv_vs_residual} and Section~\ref{sssec:ent_model_reduct}.

\subsubsection{Prediction Conviction}
\label{sec:prediction_conviction}

We define \emph{prediction conviction} is the amount of surprisal required to predict a value given a model given the model's uncertainty.  To characterize the model's uncertainty, we use residuals.

\begin{definition}
  Let $\xi$ be the number of features in a model and $n$ the number of observations. We define the \emph{residual function}, $r: X \to \mathbb{R}^{\xi}$, on the training data $X$ as
  \begin{equation}
    r(x) = J_1^\Omega(x), J_2^\Omega(x), \dots, J_\xi^\Omega(x),
    \label{eq:residual_definition}
  \end{equation}
  where $J_i^\Omega$ is the residual of the model on feature $i$ at point $x$, parameterized by a set of hyperparameters $\Omega$. We will refer to the residual function evaluated on all of the model data as $r_M$.
\end{definition}

Typically, the feature residuals will be calculated as mean absolute error or standard deviation.  Further, subsets of features may be used to compute the residual, particularly when performing targeted operations.

\begin{definition}
  Given a point $x \in \mathbb{X}$ and the set $K$ of its $k$ nearest neighbors, a distance function $d : \mathbb{R}^z \times \mathbb{Z} \to \mathbb{R}$, and a distance exponent $\alpha$, the \emph{distance contribution} of $x$ is the harmonic mean

  \begin{equation}
    \phi(x) = \left(\frac{1}{|K|}\sum\limits_{k\in K}\frac{1}{d(x, k)^{\alpha}}\right)^{-1}.
 \label{eq:contrib}
  \end{equation}

\end{definition}

The distance contribution reflects how much ``distance'' a point contributes to a graph connecting the nearest neighbors, which is the inverse of the density of points over a unit of distance in the Lebesgue space.  The harmonic mean of the distance contribution reflects the inverse of the inverse distance weighting often employed with kNN, though other techniques may be substituted if inverse distance weighting is not employed.

We can quantify the information needed to express a distance contribution $\phi(x)$ by transforming it into a probability. We begin by selecting the exponential distribution to describe the distribution of residuals as it is the maximum entropy distribution constrained by the first moment.\footnote{Other distributions may be selected by adjusting the assumptions slightly, such as the log-normal distribution.  The log-normal distribution is the maximum entropy distribution assuming that we know the standard deviation rather than the mean, but this distribution assumes that something closer is less likely, and may be better suited for familiarity conviction.  Further, the exact distribution of the distance contribution may be solved if the distributions of the features are known.}  We represent this in typical nomenclature for the exponential distribution using the norm from Equation~\ref{eq:frac_norm} as
\begin{equation}
  \frac{1}{\lambda} = ||r(x)||_p.
\end{equation}

We can directly compare the distance contribution and p-normed magnitude of the residual.  This is because the distance contribution and the norm of the residual are both on the same scale, with the distance contribution being the expected distance of new information that the point adds to the model, and the norm of the residual is the expected distance of deviation.  Given the entropy maximizing assumption of the exponential distribution of the distances, we can then determine the probability that a distance contribution is greater than or equal to the magnitude of the residual $||r(x)||_p$ in the form of cumulative residual entropy \citep{rao2004cumulative} as
\begin{equation}
P(\phi(x) \geq ||r(x)||_p) = e^{-\frac{1}{||r(x)||_p} \cdot \phi(x)}.
\end{equation}

We then convert the probability to self-information as
\begin{equation}
I(x) = -\ln P(\phi(x) \geq ||r(x)||_p),
\end{equation}
which simplifies to
\begin{equation}
\label{eq:prediction_conviction_self_info}
I(x) = \frac{\phi(x)}{||r(x)||_p}.
\end{equation}

As the distance contribution decreases, or as the residual vector magnitude increases, less information is needed to represent this point. We can then compare this to the expected value in regular conviction form, yielding a prediction conviction of
\begin{equation}
\pi_p = \frac{\mathbb{E} I}{I(x)},
\label{eq:prediction_conviction}
\end{equation}
where $I$ is the self-information calculated for each point in the model.

\subsubsection{Prediction Contribution}		
\emph{Feature prediction contribution} is motivated by mean decrease in accuracy (MDA) \citep{archer2008empirical}. In MDA, scores are established for models with all the features, $M$, and models with each feature held out, $M_{-f_i}, i = 1 \dots \xi$. The difference $|M-M_{f_i}|$ is the importance of each feature, where the result's sign is altered depending on whether the goal is to maximize or minimize score.  Feature prediction contribution differs from MDA in that feature prediction contribution measures the conditional entropy of adding a feature.  This means using prediction conviction on features with significant information may yield higher contribution values even if the feature is independent.

Prediction contribution information, $\pi_c$, is correlated with accuracy and thus can be used as a surrogate.  The expected self-information required to express a feature is given by
\begin{equation*}
  \mathbb{E} I(M) = \frac{1}{\xi} \sum_{i = 0}^\xi I(x_i),
\end{equation*}
and the expected self-information to express a feature without feature $i$ is
\begin{equation*}
  \mathbb{E} I(M_{-i}) = \frac{1}{\xi} \sum_{j = 0}^\xi I_{-i}(x_j).
\end{equation*}
From these equations, we can more formally define prediction contribution of a feature and prediction conviction of a feature.

\begin{definition}
The \emph{prediction contribution of a feature}, $\pi_c$, of feature $i$ is
\begin{equation*}
  \pi_c(i) = \frac{\mathbb{E} I(M) - \mathbb{E} I(M_{-f_i}) }{\mathbb{E} I(M)}.
\end{equation*}
\end{definition}

\begin{definition}
The \emph{prediction conviction of a feature}, $\pi_p$, of feature $i$ is
\begin{equation*}
  \pi_p(i) = \frac{ \frac{1}{\xi} \sum_{i = 0}^{\xi} \mathbb{E} I(M_{-f_i}) }{ \mathbb{E} I(M_{-f_i}) }.
\end{equation*}
\end{definition}

\subsubsection{Familiarity Conviction}

\emph{Familiarity conviction} is a metric for describing surprisal of points in a model relative to the training data.  This differs fundamentally from prediction conviction.  Consider a data set that has data points at regular intervals, such as a data point for each corner in a grid.  Given this grid data, prediction conviction will indicate that a data point very close to an existing data point will not be surprising and that it should be easy to predict given the low level of uncertainty.  However, familiarity conviction would indicate a higher surprisal for such a point even though it is easy to label because the point is unusual with regard to the even distribution of the rest of the data points.  This new point does not form another corner of the grid.  The pair of prediction conviction and familiarity prediction can be used together to find and remove data points that are easy to predict but unusual with regard to uniqueness of data.  These properties make familiarity conviction valuable for sanitizing data and reducing data as well as extracting patterns and anomalies, as is discussed in other sections.

Familiarity conviction is based on the similarity metrics as described in Section~\ref{ssec:sim_metric}.  As long as a low or zero value of $p$ is used in $L_p$ space metrics for similarity, familiarity conviction is independent of the scale of the data and provided and does not overreact to feature dominance based on feature scale and range.

\begin{definition}
  Given a set of points $X \subset \mathbb{R}^z$ for every $x\in \mathbb{X}$ and an integer $1\leq k < |X|$ we define the \emph{distance contribution probability distribution}, $C$ of $X$ to be the set
  \begin{equation}
   C = \left\{\frac{\phi(x_1)}{\sum_{i=1}^{n}\phi(x_i)}, \frac{\phi(x_2)}{\sum_{i=1}^{n}\phi(x_i)}, \dots, \frac{\phi(x_n)}{\sum_{i=1}^{n}\phi(x_i)}\right\}
  \label{eq:distance_probs}
  \end{equation}
for a function $\phi:X \to \mathbb{R}$ that returns the distance contribution.
\end{definition}

Note that because $\phi(0) = \infty$ may be true under some circumstances, multiple identical points may need special consideration, such as splitting the distance contribution among those points.

\begin{remark}
    Clearly $C$ is a valid probability distribution. We will use this fact to compute the amount of information in $C$.
\end{remark}

\begin{definition}
  The \emph{point probability} of a point $x_{i}, i=1,2,\dots,n$ is
  \begin{equation}
    l(i) = \frac{\phi(x_i)}{\sum\limits_{i}\phi(x_i)},
    \label{eq:pprob}
  \end{equation}
  where we see the index $i$ is assigned the probability of the indexed point's distance contribution.
\end{definition}

\begin{definition}
 We the set of random variables that characterize the \emph{discrete distribution of point probabilities}, $L$, is the set of $L = \{l(1), l(2), \dots, l(n)\}$.
\end{definition}

\begin{remark}
  Because we have no additional knowledge of the distribution of points other than they follow the distribution of the data, we assume $L$ is uniform as the distance probabilities have no trend or correlation.
\end{remark}

\begin{remark}
  A distance contribution is a discrete distribution of point probabilities.
\end{remark}

\begin{definition}
  The \emph{familiarity conviction} of a point $x_i\in X$ is

  \begin{equation}
    \pi_f(x_i) = \frac{ \frac{1}{|X|} \sum\limits_{i} \mathbb{KL}\left(L || L - \{i\}\cup \mathbb{E} l(i) \right)}{ \mathbb{KL}\left(L || L - \{x_i\}\cup \mathbb{E} l(i) \right) },
    \label{eq:familiarity_conviction}
  \end{equation}

  where $\mathbb{KL}$ is the Kullback-Leibler divergence. Since we assume $L$ is uniform, we have that the expected probability $\mathbb{E} l(i) = \frac{1}{n}$.
\end{definition}

Equation~\ref{eq:familiarity_conviction} can thus be used to compute familiarity conviction.

\iffalse
\subsubsection{Conviction Versus Residuals}
\label{sssec:conv_vs_residual}

-high conviction with large residuals, etc.
\fi

\subsection{Consequences of kNN Endowed with Conviction}

Having defined two methods (convictions) to measure surprisal in the space of points $X$, we introduce techniques that naturally fall out of the information-based representation. The ability to add and remove points to a model easily without retraining, coupled with familiarity and prediction conviction, enable numerous applications in model compression, online learning, anomaly detection, model to model comparison, reinforcement learning, synthetic data generation, and likely more techniques we have not considered. We detail some of these in the subsections following.

\subsubsection{Entropy-Based Model Reduction}
\label{sssec:ent_model_reduct}

Accuracy is best from models with more data, but using kNN with additional data comes at a computational cost.  Even for models where that relationship does not hold, the memory needed to store the data can be expensive.  This important problem has received some attention with some promising results despite the difficulty of being an NP-hard problem \citep{gottlieb2014near,kontorovich2017nearest}.  Our probability and entropy based approach offers some new ways of looking at this problem in an interpretable manner, relaxing some constraints with regard to strict metric spaces.

The entropy measures we discuss can be employed for pruning the model of cases as well as for targetless feature pruning, thereby reducing data size while retaining information in the model.  Overall, performing feature or case pruning can benefit any system by reducing the memory and possibly computational resources needed to use the model.  Further, these techniques can be used to help direct training, that is, which parts of the model may benefit from having more data.  We note that model reduction may be targetless, and in such cases is not a substitute for feature engineering for a targeted application.  Rather, it is useful for removing redundant data to focus on the most or least surprising relationships.

Our entropy-based techniques are generally applicable to either feature or case pruning based on the amount of information that each feature or case provides to the model.  A first step in data reduction can be to detect and remove anomalies from the model.  Anomalous cases, arguably, reduce the average usefulness of decisions the model makes.  Therefore, removing anomalies can actually improve the usefulness of the model while slightly reducing the model size.  A next step to reduce model size would be to use our techniques to remove those cases or features that do not provide significant-enough amounts of information to the model.

There are a few approaches we use to determine which cases or features to prune.  We can look at the surprisal of the case relative to the rest of the model.  If the surprisal is low, the case may be removed from the model for being redundant, which may be done dynamically for online learning.  In some circumstances, we keep the instances with the highest surprisal (i.e., the most informative) in order to cap the model size, or remove the case with the lowest surprisal in order to reduce the model size by a set amount.  We can also keep the cases with a surprisal above a certain threshold, and perhaps vary that threshold, balancing model size and information.  This also applies to features in the model.  Those features with high surprisal can be kept (or alternatively, those features with low surprisal can be removed), thereby reducing the model size in the feature space.  Further, both case pruning and feature pruning could be performed, reducing the model in both dimensions.

Though we have outlined a few surprisal and conviction measures in this paper for model reduction, they can be combined other ways to perform model reduction, or used with other feature engineering techniques.  For example, feature prediction conviction could be used to determine whether to include features in a targetless model, based on high or low surprisal for predictions.  Feature familiarity conviction could be used to determine which features have training data that most conforms to typical training data reflected in the rest of the model.

\subsubsection{Model-to-Model Surprisal}

In that kNN models are data, we can compare them using prediction and familiarity conviction directly. The use of such a comparison is of particular interest in online learning where one could detect a non-stationary process by comparing models built during different times.  In reinforcement learning one could compare models to establish propensity for exploration verses exploitation.

Further, by measuring the amount of surprisal that one model contributes to another, some sparse machine learning problems can be transformed into dense problems.  Consider numerous individuals that occasionally take some action of a certain type.  A model may be trained for each individual, and then compared with surprisal against other individuals or against larger aggregated models to find which larger model would be least surprising for the individual to be included in.  We have found promising initial results in commercial settings with these techniques.  Combining these untargeted techniques with targeted model selection techniques, such as Bayesian information criterion \citep{schwarz1978estimating} or Akaike information criterion \citep{akaike1973information} is another potential future direction.

\subsubsection{Data Imputation and Overcoming Data Sparsity}

\label{sssec:imputation}

Sparse data is an issue for many machine learning systems, which fail to perform or perform poorly when trained with sparse data, especially if the data is sparse across multiple features.\footnote{We are referring to the limited semisupervised learning techniques that are typically available to by data scientists.  There are a number of statistics techniques available if sufficient information about a distribution is known, and domain specific imputation techniques, particularly for language processing, such as that of \cite{gemmeke2008using}.}  The sparse data can come from sporadic unobservability (i.e., failure to capture the feature at particular times) and unavailability (e.g., no collection of certain feature at certain times, a common issue with historical datasets).  Regardless of the cause, the combination of targetless kNN with entropy can be used to overcome sparsity by imputing missing features in sparse data, and doing so in a way that reflects the dataset.

In past approaches, generically solving for missing values in a data set has presented many challenging problems, such as Boolean satisfiability and other NP-complete and NP-hard problems. When the rules that determine the missing values are also unknown, the problem becomes even more difficult. Semi-supervised learning has been a rich area of study to attempt to tackle this problem \citep{triguero2015self}. Our targetless approach, when combined with conviction, offers a novel technique for imputation that can fill in missing data across all of the features and scaffold knowledge from cases with missing data as it fills the missing data in.

The algorithm finds the cases with missing data that are least surprising, labels them, inserts the label into the model for the case, and repeats the process until no more missing data exists. Grouping the cases into batches (i.e. imputing multiple missing data points in a batch) can improve performance so that conviction only needs to be computed occasionally.

More specifically, the algorithm orders the cases with missing data by their surprisal for each feature, conditioned only on the data known for the case, from lowest to highest.  For particularly sparse data, we have had success with first ordering the cases by those with the fewest nulls prior to ordering by entropy.\footnote{Currently, we use entropy for the cases to determine what data to impute first, but we are also exploring the use of entropy for features.  We first order by feature entropy, and then by entropy of the cases missing the lowest entropy feature.  We would impute that data first, then recalculate entropies for the model, similar to the manner described above.}  Starting with the case with the lowest entropy, we determine a value for the missing data using kNN based on the known features.\footnote{We can also use the techniques described in this paper for synthetic data generation to determine the value for the missing feature as described in Section~\ref{sssec:synth_data}.}  We can determine values for multiple features at once (in batch), or determine just one missing value, and then redetermine the entropy for the cases, and return to impute more data.  This process continues until all of the data is complete.  When we want to merely reduce the sparsity of the data (vs. filling in all of the values), we can continue until we hit a termination condition or sparsity threshold.  Additional work is needed, but we believe a viable termination condition is ceasing data imputation when the entropy for the lowest-entropy case exceeds a threshold.  We believe that, after that point, the additional informational value of the imputed data may be low, and, as such, we cease imputing missing data.

\subsubsection{Auditability, Explanations, and Automated Decision Making}
\label{ssec:auditable_explanations}

The choice of a kNN model allows us number of interesting opportunities related to providing explanations of the data.  With kNN, the model is the data.  The model is only complemented by kernel parameters to define nearest neighbors, and with any weights, adjustments or removal of data.  In both the assessment of the model for decision or action suggestion, the data points that are within the kernel, or that are above a certain threshold for how much they impact the kernel values, are the data points that caused the decision. This is compelling from a machine decision auditability perspective because it means that the data relevant to each decision is directly identified, and that auditing, editing or removing the training data identified as being associated with a decision would have had a direct effect on the decision.

Below, we discuss numerous types of explanation data that can be derived from a kNN model.  For each, we give a short description of how it is generated, and what a system or operator can do with it.  As an overview, for each type of explanation data (or based on combinations of explanation data), that data can be passed to a system or human operator for review.  The system or human operator can use the explanation data to decide whether to perform the action in question, perform a different action, or perform no action at all.  Additionally, this data can be used after-the-fact to audit decisions that were made. Auditability begins, as alluded to above, with the fact that we can determine what data was used to suggested an action. The explanation data can also be generated either at the time of suggesting the action or at the time of the audit.

One of the measures we use as part of our explanation data is conviction, which can be seen, broadly, as the ratio of the amount of surprisal of a particular case to the actual surprisal. Numerous types of conviction are discussed in this paper, such as familiarity conviction and prediction conviction. We also discuss formulating each in a targeted or untargeted manner. Any of these formulations for conviction may be used and provided as an explanation data along with answers or suggested actions. As an example, targeted or untargeted familiarity conviction may be calculated and provided to along with a suggested decision. An overly low (or high) conviction score can be a cause for concern, and that suggested decision may be flagged for further reviewed or ignored completely. In contrast, a suggested decision or action with a conviction score that is not concerning (e.g., a moderate conviction score, or, in some circumstances, a high conviction score), may be performed or acted upon without further review. A low conviction score may be associated with the data being outside the usual pattern of the data, and therefore be of concern for systems that are trying to perform ``usual'' actions. High conviction may be of concern when there is a desire to not perform actions that are too ``usual.'' A high and low pass filter can be used when there is a desire to perform actions that are neither overly usual or unusual, but instead are somewhere in the middle in terms of how expected they are. 

We can also use feature prediction contribution or feature prediction contribution to find bias in decision making. Many models and data sets contain many features, and are used for many decisions. As described elsewhere herein, our techniques can help prune a model of cases and features. Nevertheless, there may always be features in models that, in an ideal world, would contribute little to nothing to certain decisions. For example, in the context of financing decisions to individuals, there are certain features that, if decisions were made based on them, would constitute undesirable bias. These may include gender or race. The feature prediction contribution can be provided to the system or human operator to help determine whether making the decision based on that feature would constitute unpermitted bias. This could lead to pruning that feature from the model or taking other steps to reduce or eliminate its impact on the decision, such as the use of feature weights.

The local region of the model (which we will refer to as the \emph{local model}) comes with no added cost with kNN and therefore we can perform analysis on the local model and make other queries on relevant data.  As a few examples, we can easily find counterfactual cases \citep{wachter2017counterfactual} and the boundary conditions they yield by performing a query on the data that maximizes the ratio of the action features to the total set of features. We can also find out if any of the features in a case associated with the actions suggested by the model are outside the range of the corresponding features of the cases in the local model.  Features being outside the range may be cause for concern and may cause the suggested action to not be performed.

As a counter-point to the counterfactual cases, we can also determine archetype cases, cases that have the same action as the suggested action, and are furthest from other cases with different actions.  The distance from the case with the suggested action to the archetype case can be used as explanation data, audit data, and to determine whether to perform a suggested action without further review.

We can find ratios of different types of conviction in the local model to those of the total model, which can indicate which cases and features add significant information to a particular region, and which add less information. If the information is not correlated with accuracy (e.g., has low conviction in the local model and high conviction in the model as a whole), then it may be a measure of noise. If the cases that contribute to the suggestion of an action are above a noisiness threshold, then it may be inadvisable to perform a suggested action. On the other hand, if the noisiness of the cases used to suggest an action is low, the system or operator may be confident in performing the action. There are many other ratios of conviction that may be used as explanation and audit data, and different of them will be useful in different scenarios. 

A ``less similar'' model can also be determined, where the closest k cases (by distance, by count, by density threshold) are excluded, and the distance to the next closest cases is determined.  That distance can be used as explanation data, audit data, and to determine whether to perform a suggested action without further review.  For example, a higher distance to the ``less similar'' cases can be an indicator that the suggested action is in a sparsely populated part of the model, and therefore should be reviewed before being acted upon. 

We can also define feature residuals based on the local model (or a regional model that is the N nearest neighbors, where N can be the same, higher, or lower than the k used to define the local model).  Here, we can use the mean absolute error, variance, or other moments or measures to predict how well the model predicts each feature when it is removed.  We can also determine action probabilities, which, in the case of categorical actions, can be measured as the percentage of cases in the local model that have that categorical action.  For continuous or ordinal actions, the action probability can be a probability measure based on the confidence interval of the suggested actions for a given tolerance (e.g., an action value for 250 may be 67\% (the action probability) likely to be within +/-5 (the tolerance) of 250).  Feature residuals and conviction can be used in conjunction; a prediction with high prediction conviction but also wide residuals may not be reliable, but a prediction with low prediction conviction but wide residuals may potentially be improved if further training data is added similar to the prediction.  We can also determine local or regional model complexity (i.e., whether the variance is high, whether the accuracy is low, whether correlations among variables are low, etc.) and fractal dimensionality (i.e., placing a shape over the model and shrinking the scale of the shape and counting the number of smaller shapes needed to cover the extents of the model).

\subsubsection{Synthetic Data Generation and Reinforcement Learning}

Given that prediction conviction is a method to express how surprising an observation is, we can reverse the math and use conviction to generate a new sample of data for a given amount of surprisal.  The general approach is to randomly select or predict a feature of a case from the training data and then resample it based on the new condition.  This approach is related to Gibbs sampling \citep{martino2015fast,efros1999texture} in that it incrementally obtains new values for each feature conditioned on the previous values, though the conditioning and sampling is based on our approach to kNN.

Using the conditioned local residuals for a part of the model, as described in section~\ref{sec:prediction_conviction}, we can parameterize the random number distribution to generate a new value for a given feature.  Our resampling method is related to the approach used by the Mann-Whitney test \citep{mann1947test}, a powerful and widely used nonparametric test to determine whether two sets of samples were drawn from the same distribution.  In the Mann-Whitney test, samples are randomly checked against one another to see which is greater, and if both sets of samples were drawn from the same distribution then the expectation is that both sets of samples should have an equal chance of having a higher value when randomly chosen samples are compared against each other.  Our approach for resampling a point is to randomly chose whether the new sample is greater or less than the other point and then draw a sample from the distribution using the feature's residual as the expected value.  Just as the exponential distribution is entropy maximizing given the sole constraint of a positive mean, the double-sided exponential distribution (also known as the Laplace distribution) is the entropy maximizing distribution given a positive mean distance about a point.  The log-normal and other distributions may be used as well, depending on the types of residuals computed and assumptions made about the local distributions.

If a feature is not continuous but rather nominal, then the local residuals can populate a confusion matrix, and an appropriate sample can be drawn based on the probabilities for drawing a new sample given the previous value.

Suppose we would like to generate synthetic data with features $i \in \Xi$. If there are no conditions placed on the new synthetic data, then we start with a random feature i.  Because the observations within the model are representative of the observations made so far, a random instance is chosen from the observations using the uniform distribution over all observations.  Then the value for feature $i$ of this observation is resampled via the methods mentioned above.  The value for feature $i$ then become a condition on subsequently-generated features.

Next, suppose that we would like to generate feature $j$, given that features $i \in \Xi$ have corresponding values $x_i$.  The model labels feature $j$ conditioned by all $x_i$ to find some value $t$.   This new value $t$ becomes the expected value for the resampling process described above, and the local residual (or confusion matrix) becomes the appropriate parameter or parameters for the expected deviation to find the value for $x_j$.

The process for filling in the features for an instance may begin with no feature values subject to conditions, or some feature values may have been specified as conditions for the data to generate.  Either way, the remaining features may be ordered (i.e., selected for determination of a new value) randomly or may be ordered via a feature conviction value.  When a new value is generated, then the process restarts with the new value as an additional condition.

\subsubsection{Parameterizing Synthetic Data Via Prediction Conviction}
\label{sssec:synth_data}

Continuing with the double-sided exponential distribution as a maximum entropy distribution of distance in $L_p$ space, we can derive a closed form solution for how to scale the exponential distributions based on a prediction conviction value.

Starting with Equation~\ref{eq:prediction_conviction}, we specify a value, $\nu$, for the prediction conviction as
\begin{equation}
	\nu = \pi_p(x) = \frac{ \mathbb{E}I }{I(x)}
\end{equation}
which can be rearranged as
\begin{equation}
	I(x) = \frac{\mathbb{E} I}{\nu}.
\end{equation}
Substituting in the self-information from Equation~\ref{eq:prediction_conviction_self_info}, we have
\begin{equation}
	\label{eq:datagen_exponential_equality}
	\frac{\phi(x)}{||r(x)||_p} = \frac{\mathbb{E} I}{\nu}.
\end{equation}
Note that the units on both sides of Equation~\ref{eq:datagen_exponential_equality} match.  This is because of the natural logarithm and exponential in the derivation of Equation~\ref{eq:datagen_exponential_equality} cancel out, but leave the resultant in nats.  We can rearrange in terms of distance contribution as
\begin{equation}
	\label{eq:datagen_dist_contrib}
	\phi(x) = \frac{ ||r(x)||_p \cdot \mathbb{E} I}{\nu}.
\end{equation}

To proceed further we need to make an assumption about the distribution of distance contributions $\phi$. Seeking to minimize the complexity of our assumptions we simply observe that distances are supported by the positive reals. Constraining the first or first and second moments and maximizing the entropy gives us the exponential and log normal distributions respectively. For simplicity sake we proceed with $e(\zeta)$ but note that in practice $\ln \mathcal{N}(\mu, \sigma)$ is often observed. One may distinguish among the distributions using likelihood curvature tools such as Fisher Information.

If we let $p = 0$, which is desirable for conviction and other aspects of the similarity measure, we can rewrite the distance contribution in terms of a norm of the values observed for the number of features, $\xi$, each with an expected mean of $\frac{1}{\zeta_i}$. Taking the expected value of both sides we find
\begin{equation}
	\left(\Pi_i \frac{1}{\zeta_i} \right)^\frac{1}{\xi} = \frac{ \left(\Pi_i r_i\right)^\frac{1}{\xi} \mathbb{E} I}{\nu}.
\end{equation}
Due to the number of ways of assigning surprisal across the features, many solutions may exist.  However, unless otherwise specified or conditioned, we would want to distribute surprisal across the features holding expected proportionality constant.  This allows us to write the distance contribution, which becomes the mean absolute error for the exponential distribution, as
\begin{equation}
	1 / \zeta_i = r_i \left(\frac{\mathbb{E} I}{\nu}\right)^\xi.
\end{equation}
and solving for the $\zeta_i$ to parameterize the exponential distributions, we find
\begin{equation}
	\label{eq:datagen_exponential_parameter}
	\zeta_i = \frac{1}{r_i} \left(\frac{\nu}{\mathbb{E} I}\right)^\xi.
\end{equation}
Equation~\ref{eq:datagen_exponential_parameter}, taken with the value of the feature, becomes the distribution by which to generate a new random number under the maximum entropy assumption of exponentially distributed distance from the value.

\subsubsection{Reinforcement Learning}

The ability to randomly generate data with a controlled amount of surprisal is a novel way to characterize the classic exploration versus exploitation trade off in searching for an optimal solution to a goal.  Currently, pairing a means to search, such as Monte Carlo tree search \citep{abramson1987mcts}, with a universal function approximator, such as neural networks, is the most successful approach to solving difficult reinforcement learning problems without domain knowledge \citep{silver2017mastering}.  Because our data synthesis technique comes from the universal function approximator model (kNN) itself, we can create a reinforcement learning architecture that is similar and tightly coupled.

Because the synthetic data generation can be conditioned, we can condition the search on both the current state of the system, as it is currently observed, and a set of goal values for features.  As the system is being trained, it can be continuously updated with the new training data.  Once states are evaluated for their ultimate outcome, a new set of features or feature values can be updated or added to all of the observations indicating the final scores or measures of outcomes.  Keeping track of which observations belong to which training sessions (or games) is a convenient way to track and update this data.  Given that the final score or multiple goal metrics are already in the kNN database, the synthetic data generation can query for new data conditioned upon having a high score or winning conditions, with a specified amount of conviction.

This results in a reinforcement learning algorithm that can be queried for the relevant training data for every decision, as described in Section~\ref{ssec:auditable_explanations}.  The commonality among the similar cases, boundary cases, archetypes, etc.\ can be combined to find when certain decisions are likely to yield a positive outcome, negative outcome, or a larger amount of surprisal thus improving the quality of the model.  By seeking high surprisal moves, the system will improve the breadth of its observations and learning, though it may not perform well.  Setting the conviction of the data synthesis to $1$ yields a balanced trade off between exploration versus exploitation.  As more information is learned, this conviction value may be reduced to focus on achieving goals.

The interpretability of reinforcement learning may help overcome many of the data-availability issues.  For example, when data is needed for dangerous, expensive, or otherwise difficult-to-produce training data, we can generate synthetic data conditioned in value and conviction to match those difficult circumstances.  As such, our method can provide the sampling strategy necessary for reinforcement learning with more control than with current techniques.

\section{Initial Results and Future Direction}

Although providing a rigorous review of our results and methods is out of the scope of this paper, we summarize a few here to motivate and encourage additional exploration.  We tested classic kNN as implemented in scikit-learn \citep{pedregosa2011scikitlearn}, standardizing the scale of the features as is common practice in machine learning, against kNN with fractional p-values and $\lim_{p\to 0}$ using uncertainty in distance as mentioned in Section~\ref{sssec:probabilistic_distance} \emph{without} standardization.  We compared the results across a robust suite of 97 regression datasets and 78 classification datasets selected from among the benchmark data published by \cite{olson2017pmlb}.

On the classification datasets, scikit-learn's random forest implementation averaged an accuracy of 0.79.  kNN already performs well on classification problems, averaging an accuracy of 0.76.  However, using fractional p values, we saw the accuracy increase to 0.77, and allowing the p value of 0 based on uncertainty we saw the average accuracy improve to 0.78.  Though the accuracy improvements of our techniques are slight, the use of low or zero p values means that we can maintain the data directly without scaling and that we have made a step toward accurate probability-based reasoning on data using conjunctions as described in Section~\ref{sssec:probabilistic_distance}.

On the regression datasets, scikit-learn's random forest implementation averaged an r-squared score of 0.77.  In many situations, such as those involving extrapolation, kNN regression does not perform as well as other methods, and we saw this with the scikit-learn implementation resulting in an r-squared score of 0.53.  However, our improvements yielded considerable gain in kNN's regression scores.  Using fractional p values yielded an r-squared score of 0.57, which is a significant (p $\ll$ .001) improvement based on the Wilcoxon signed rank test.  Further allowing the use of a zero p value with uncertainty in distance as mentioned in Section~\ref{sssec:probabilistic_distance}, the r-squared score improved to 0.66 which is also a significant result (p $\ll$ .001).

We believe that kNN's regression scores can also be improved to be competitive with other cutting edge algorithms.  In Section~\ref{sssec:imputation}, we showed how a targetless approach to data can fill in missing data, and from an auditability perspective it is easy to track the history of data imputation.  Conversely, we are investigating \emph{exputation} approaches can be employed to synthesize likely data points outside the bounds of the training data.  Knowledge of the features, such as their bounds, can help when reflecting or amplifying or synthesizing exputed data points, which can then be used for interpolation.

The reason for our belief that this is a core problem lies with the topology of data as the dimensionality grows.  As the number of dimensions increases for a given set of data, many intuitive analytical techniques such as Euclidian norms and Gaussian kernels become inappropriate as the unit radius hypervolume goes toward zero and the probability that data points falling in sharp corners of a hypervolume goes toward one~\citep{verleysen2005curse}.  This implies that nearly all data points will be at or beyond the periphery, requiring extrapolation.  Dealing with any kind of cost or value function to perform optimization will mean that nearly all points are Pareto optimal, meaning that it becomes increasingly more difficult to define ``good'' because nearly every point has some unique quality.  In many cases, the larger number of dimensions can be helpful, but primarily when the structure is extracted and the dimensionality is reduced~\citep{KittlerFeatureSelection, KohaviWrappers, StoppigliaRanking}.  As nearly all new observations will be on the periphery, we believe extrapolation techniques, such as exputation, are likely to improve results while maintaining interpretability.

A dimensionality bottleneck is little different than the information bottlenecks used for generalization and variance reduction across other areas of machine learning~\citep{tishby2015deep}.  By employing clustering techniques in conjunction with model reduction techniques as mentioned in Section~\ref{sssec:ent_model_reduct}, selecting prototypes or archetypes to represent a cluster in a hierarchical fashion, we may be able to characterize the entropy flux between parts of the model, and hierarchical models and hierarchical explanations are natural consequences.  We note the striking commonality of zero p value Lebesgue space as depicted in Appendix~\ref{ap:geo_mean_derrive}, the conjunction of probabilities of independent distributions, and the core of the no-flattening theorem by \cite{lin2017does} that relates hierarchical architectures for neural networks and performance.  Our future work will include further efforts to use probability throughout all parts of kNN such that any form of entropy or probability can be calculated, and assumptions can be clearly interpreted.

Additional future work will be to characterize our work in the performance and scalability of targetless kNN queries with fractional and zero p values, which is outside the scope of this paper.

Maximizing the interpretability of artificial intelligence leads to either understanding the generalized relationships of the data, such as symbolic or tree-based models, or to understand the data itself.  With the improved performance of computing and the advances in kNN, we conclude that using kNN provides a promising foundation for the future of interpretable artificial intelligence and machine learning.

\bibliographystyle{abbrvnat}
\bibliography{swiftrain.bib}

\appendix
\section{Geometric Mean Derivation}
\label{ap:geo_mean_derrive}

The geometric mean can be derived from the generalized mean as
\begin{align*}
\lim_{p \to 0} \left(\sum_{i=1}^n w_i x_{i}^p \right)^{1/p} &= \lim_{p \to 0} \exp{\left( \ln{\left[\left(\sum_{i=1}^n w_i x_{i}^p \right)^{1/p}\right]} \right) }\\
&= \lim_{p \to 0} \exp{\left( \frac{\ln{\left(\sum_{i=1}^n w_i x_{i}^p \right)}}{p} \right) }.
\end{align*}

Then using L'H\^opital's rule and the chain rule on the inner part of this equation, we can simplify as
\begin{align*}
\lim_{p \to 0} \frac{\ln{\left(\sum_{i=1}^n w_i x_{i}^p \right)}}{p} &= \lim_{p \to 0} \frac{\frac{\sum_{i=1}^n w_i x_i^p \ln{x_i}}{\sum_{i=1}^n w_i x_i^p}}{1} \\
&= \lim_{p \to 0} \frac{\sum_{i=1}^n w_i x_i^p \ln{x_i}}{\sum_{i=1}^n w_i x_i^p}\\
&= \frac{\sum_{i=1}^n w_i \ln{x_i}}{ {\sum_{i=1}^n w_i} }\\
&= \frac{\ln{\left(\prod_{i=1}^n x_i^{w_i} \right)}}{{\sum_{i=1}^n w_i}}.
\end{align*}

Therefore substituting back in the previous result yields
\begin{align*}
\lim_{p \to 0} \left(\sum_{i=1}^n w_i x_{i}^p \right)^{1/p} &= \lim_{p \to 0} \exp\left( \frac{ \ln{\left(\prod_{i=1}^n x_i^{w_i} \right)} }{\sum_{i=1}^n w_i} \right)\\
&= \left(e^{\left( \ln{\left(\prod_{i=1}^n x_i^{w_i} \right)} \right)} \right)^{\left( \frac{1}{\sum_{i=1}^n w_i} \right)}\\
&= \left(\prod_{i=1}^n x_i^{w_i}\right)^{\left( \frac{1}{\sum_{i=1}^n w_i} \right)}.
\end{align*}

Setting all $w_i = \frac{1}{n}$ yields
\begin{equation*}
\lim_{p \to 0} \left(\sum_{i=1}^n \frac{1}{n} x_{i}^p \right)^{1/p} = \left(\prod_{i=1}^n x_i \right)^{\frac{1}{n}}.
\end{equation*}

\end{document}